\documentclass{article}
\usepackage{cite}
\usepackage{amsmath,amssymb,amsfonts}
\usepackage{algorithmicx}
\usepackage[ruled]{algorithm2e}
\usepackage{graphicx}
\usepackage{subfigure}
\usepackage{epstopdf}
\usepackage{caption}
\usepackage{textcomp}
\usepackage{booktabs}
\usepackage{multirow}

\title{Binary Matrix Completion with Nonconvex Regularizers}
\author{Chunsheng Liu, Hong Shan}

\begin{document}
	\maketitle
\begin{abstract}
Many practical problems involve the recovery of a binary matrix from partial information, which makes the binary matrix completion (BMC) technique received increasing attention in machine learning. In particular, we consider a special case of BMC problem, in which only a subset of positive elements can be observed. In recent years, convex regularization based methods are the mainstream approaches for this task. However, the applications of nonconvex surrogates in standard matrix completion have demonstrated better empirical performance. Accordingly, we propose a novel BMC model with nonconvex regularizers and provide the recovery guarantee for the model. Furthermore, for solving the resultant nonconvex optimization problem, we improve the popular proximal algorithm with acceleration strategies. It can be guaranteed that the convergence rate of the algorithm is in the order of ${{1 \mathord {\left/{\vphantom {1 T}} \right.\kern-\nulldelimiterspace} T}}$, where $T$ is the number of iterations. Extensive experiments conducted on both synthetic and real-world data sets demonstrate the superiority of the proposed approach over other competing methods.
\end{abstract}

Binary matrix completion, Link prediction, Nonconvex regularizers, Topology inference

\section{Introduction}
\label{sec:introduction}
The matrix completion problem attempts to recover a low-rank or an approximate low-rank matrix by observing only partial elements \cite{b1}. In recent years, many strong theoretical analyses have been developed on the matrix completion problem \cite{b2,b3,b4,b5,b6,b7}, which has been applied to a wide variety of practical applications such as background modeling \cite{b8,b9}, recommender systems \cite{b10}, sensor localization \cite{b11,b12}, image and video processing \cite{b13,b14}, and link prediction \cite{b15}. In particular, these all results are based on a potential assumption that the observed entries are continuous-valued. However, in many practical applications, the observations are not only incomplete but also are often highly quantized to a single bit \cite{b16}. Therefore, there is a conspicuous gap between those existing approaches and practical situation, which promotes the rapid development of 1-bit matrix completion \cite{b16}.

Instead of observing a subset of full entries, a more common situation in practice is that only the subset of positive elements can be observed. Thus, the observations are not only binary but also nonnegative. For instance, consider the link prediction problem in social networks, where only positive relationships, such as ``friendships'', can be observed, while no ``non-friendships'' are observed. The goal here is to recover the whole social network from the observed friendships (positive entries). In the context of binary classification, the problems learned from positive and unlabeled examples are called positive and unlabeled learning (PU learning for short) \cite{b17}. Consequently, the unobserved entries were regarded as unlabeled samples, and then PU learning is applied to matrix completion \cite{b18}. 

The existing methods of PU matrix completion \cite{b18,b19} are all based on the convex regularizers such as nuclear norm and max-norm. However, many works \cite{b9,b13,b14,b20} stated that the (convex) nuclear norm might not be a good enough approximation of the rank function. In contrast, better recovery performance can be achieved by nonconvex surrogates \cite{b21,b22,b23}. Accordingly, we attempt to introduce the nonconvex regularizers into PU matrix completion. 

In this paper, we propose a novel model of PU matrix completion with nonconvex regularizers and provide recovery guarantee for the model. To cope with the challenges of the resultant nonconvex optimization problem, we improve the proximal algorithm with two acceleration schemes: i) Instead of full singular value decomposition (SVD), only a few leading singular values are needed to generate the next iteration. ii) We replace a large matrix by its projection on leading subspace, and then the reduction of matrix size makes the calculation of proximal operator more efficient. Moreover, we show that further acceleration is available by taking advantage of the sparsity structure. Subsequently, the resultant algorithm, named ``PU matrix completion with nonconvex regularizers (PUMC\_N),'' is analyzed in detail from the aspects of convergence and time complexity, respectively.

The primary contributions of our work can be summarized as follows:

\begin{itemize}
\item \emph{Employing the nonconvex regularization, we propose a novel PU matrix completion model and provide a strong guarantee for matrix recovery, i.e., the error in recovering an $m \times n$ 0-1 matrix is ${\cal O}\left( {\frac{1}{{{\delta ^2}\sqrt {mn} }}} \right)$, where $\delta$ denotes the sampling rate of positive entries.}

\item \emph{We develop an accelerated version of the proximal algorithm for solving the resultant nonconvex optimization model. It can be guaranteed that the proposed algorithm has a convergence rate of ${\cal O}\left( {{1 \mathord{\left/
				{\vphantom {1 T}} \right.
				\kern-\nulldelimiterspace} T}} \right)$, where $T$ denotes the number of iterations.}
			
\item \emph{We implement and analyze the proposed algorithm on both synthetic and real-world data sets. Experimental results demonstrate the superiority of the resultant algorithm to state-of-the-art methods.}
\end{itemize}

The paper is organized as follows. The following section is a brief overview of the related work. In Section III, we propose the model and provide its recovery guarantee. A fast and efficient algorithm is proposed in Section IV, followed by the convergence and time complexity analysis. Experimental results on both synthetic and real-world data sets are presented in Section V. Finally, the conclusion is summarized in Section VI.

\textbf{Notation}: In this paper, vectors and matrices are denoted by lowercase and uppercase boldface, respectively. For a matrix ${\mathbf{X}}$, ${{\mathbf{X}}^{ \top }}$ denotes its transpose, ${{\mathbf{X}}_k} = {\mathbf{X}}\left( {:,1:k} \right)$ is its leading $k$ columns, ${\left\| {\mathbf{X}} \right\|_F} = \sqrt {\sum\nolimits_{i,j} {X_{ij}^2} }$ is the Frobenius norm of ${\mathbf{X}}$, and ${\left\| {\mathbf{X}} \right\|_*} = \sum\nolimits_i {{\sigma _i}\left( {\mathbf{X}} \right)}$ is the nuclear norm, where ${\sigma _i}\left( {\mathbf{X}} \right)$ is the $i$-th largest singular value of ${\mathbf{X}}$. For a set $\Omega$, $\left| \Omega  \right|$ is its cardinal number. In addition, we use $\nabla f$ for the gradient of a differentiable function $f$.

\section{Related work}
In the last decade, based on the remarkable result of low-rank matrix completion \cite{b1}, a tremendous amount of work has focused on the problem, which enabled a burst of progress concerning the matrix completion theory. A strong theoretical basis of matrix completion \cite{b2,b3,b5}, including the case of approximate low-rank matrices and noisy observations \cite{b4,b1,b9,b11}, has been established.

However, these all results are based on the underlying assumption that the observed entries are continuous-valued. In practice, many applications, such as the popular \textit{Netflix}\footnote{https://netflixprize.com/index.html} and \textit{MovieLens} \cite{b16} in recommender systems, have a rating matrix whose entries are discrete and quantized rather than continuous. Consequently, there is a conspicuous gap between standard matrix completion theory and practice, revealing the inadequacy of the corresponding methods in dealing with the above case.

Motivated by the above challenge, 1-bit matrix completion was advocated for the first time in \cite{b16} to deal with the binary (1-bit) observations. Theoretical guarantees were provided to show the efficiency of the method. In addition, a suite of experiments on both synthetic and real-world data sets illustrated some of the practice applications and demonstrated the superiority of 1-bit matrix completion. Then, \cite{b24} considered a general nonuniform sampling distribution concerning 1-bit matrix completion problem followed by corresponding theoretical guarantees. Moreover, the noisy version was studied in \cite{b25} under the same sampling scheme with \cite{b24}. Instead of nuclear norm, \cite{b25} used the max-norm as a convex relaxation for rank function. Similarly, \cite{b26} addressed the problem of social trust prediction with a 1-bit max-norm constrained formulation. Under constraints on infinity norm and exact rank, the noisy 1-bit matrix completion problem was explored in \cite{b27} and \cite{b28}. Furthermore, though the analysis on PAC-Bayesian bounds, \cite{b29} evaluated the performance of 1-bit matrix completion. 

Relative to the settings of 1-bit matrix completion, there is a more common situation in practice. Consider the link prediction problem in social networks, instead of observing a subset of full entries, we can only observe a subset of the positive relationships, which is ``one-sided'' sampling in \cite{b18}. The similar situation also occurs in network topology inference problem \cite{b48}. In response to such case, \cite{b18} proposed PU matrix completion. This method introduced the idea of PU (positive and unlabeled) learning \cite{b17,b30}, i.e., learning only in the presence of positive and unlabeled examples. Motivated by the development of semi-supervised classification \cite{b31} in recent years, \cite{b19} proposed the modified version of PU matrix completion.

In particular, the PU matrix completion was considered under the constraints on the nuclear norm. As the tightest convex lower bound of the matrix rank function, the nuclear norm is the most popular convex regularizer. Many algorithms based on the nuclear norm, such as accelerated inexact soft-impute algorithm (AIS-Impute) \cite{b32}, singular value thresholding (SVT) \cite{b33}, and inexact augmented Lagrange multipliers (IALM) \cite{b34}, can solve the corresponding convex optimization problem effectively. Despite the nuclear norm is applied successfully and makes low-rank optimization easier, numerous attempts have recently been made to regard nonconvex regularizers as the better approximation of the matrix rank. For instance, nonconvex surrogates including truncated nuclear norm (TNN) \cite{b9,b35}, log-sum penalty (LSP) \cite{b22,b36}, and capped $\ell_1$ penalty \cite{b21} have been successfully applied in many fields and have better empirical performance than nuclear norm regularizers.

\section{Problem formulation}
Matrix completion is the problem of recovering the underlying target matrix given its partial information \cite{b1}. Following the ``basic setting'' of \cite{b18}, let the target matrix ${\mathbf{M}} \in {\left\{ {0,1} \right\}^{m \times n}}$ be a binary matrix that consists only of ones and zeros, and ${\Omega _1} = \left\{ {\left. {\left( {i,j} \right)} \right|{M_{ij}} = 1} \right\}$ denotes the index set of all positive elements in ${\mathbf{M}}$. Equivalently, the observation matrix is denoted, herein, by ${\mathbf{A}} \in {\left\{ {0,1} \right\}^{m \times n}}$, and $\Omega$ denotes the index set of observation elements. According to the ``one-sided'' sampling in \cite{b18}, only a subset of positive entries of ${\mathbf{M}}$ can be observed, that is $\Omega  \subseteq {\Omega _1}$. We suppose that the observation process follows the uniform sampling distribution, which is the popular choice for majority work, i.e., $\Omega$ is sampled randomly from ${\Omega _1}$. For the observation matrix ${\mathbf{A}}$, ${A_{ij}} = 1$ if $\left( {i,j} \right) \in \Omega$ and ${A_{ij}} = 0$ otherwise. Here, our goal is to recover the underlying target matrix ${\mathbf{M}}$ from the observation matrix ${\mathbf{A}}$.

According to the above description, the relationship between ${\mathbf{M}}$ and ${\mathbf{A}}$ can be expressed in the following conditional probability.
\begin{equation}
\begin{array}{l}
P\left( {\left. {{A_{ij}} = 0} \right|{M_{ij}} = 1} \right) = 1 - \delta \\
P\left( {\left. {{A_{ij}} = 1} \right|{M_{ij}} = 0} \right) = 0
\end{array}.\label{eq}
\end{equation}
where $\delta  = {{\left| \Omega  \right|} \mathord{\left/
		{\vphantom {{\left| \Omega  \right|} {\left| {{\Omega _1}} \right|}}} \right.
		\kern-\nulldelimiterspace} {\left| {{\Omega _1}} \right|}}$ 
denotes the sampling rate. We consider the problem of positive and unlabeled matrix completion (PU matrix completion) with nonconvex regularizers as the following form.
\begin{equation}
\mathop {\min }\limits_{{\mathbf{X}} \in {{\left\{ {0,1} \right\}}^{m \times n}}}  F\left( {\mathbf{X}} \right) \equiv \ell \left( {\mathbf{X}} \right) + \lambda {r_n}\left( {\mathbf{X}} \right).\label{eq2}
\end{equation}
where $\lambda$ is a regularization parameter, $\ell$ is a smooth loss function, ${r_n}$ is a nonconvex regularizer in Table \ref{table1}. In addition, \eqref{eq2} has the following characteristics.
\begin{itemize}
	\item \emph{$\ell$ is differentiable with $\beta$-Lipschitz continuous gradient, that is, it follows ${\left\| {\nabla \ell \left( {{{\mathbf{X}}_1}} \right) - \nabla \ell \left( {{{\mathbf{X}}_2}} \right)} \right\|_F} \le \beta {\left\| {{{\mathbf{X}}_1} - {{\mathbf{X}}_2}} \right\|_F}$, $\beta  > 0$. Moreover, $\ell$ is bounded from below, i.e., $\inf \ell  >  - \infty $.}
	\item \emph{${r_n}\left( {\mathbf{X}} \right) = \sum\nolimits_{i = 1}^m {r\left( {{\sigma _i}\left( {\mathbf{X}} \right)} \right)}$ is a nonconvex and nonsmooth function, where $r$ is a nondecreasing concave function and $r\left( 0 \right) = 0$.}
	
	\item \emph{${r_n}$ can be formulated as the difference of two convex functions \cite{b47}, i.e., ${r_n}\left( {\mathbf{X}} \right) = {\mathord{\buildrel{\hbox{$\scriptscriptstyle\frown$}} 
				\over r} _n}\left( {\mathbf{X}} \right) - {\mathord{\buildrel{\hbox{$\scriptscriptstyle\smile$}} 
				\over r} _n}\left( {\mathbf{X}} \right)$  , where ${\mathord{\buildrel{\hbox{$\scriptscriptstyle\frown$}} 
				\over r} _n}$ and ${\mathord{\buildrel{\hbox{$\scriptscriptstyle\smile$}} 
				\over r} _n}$ are convex. (The corresponding convex functions of the nonconvex regularizers mentioned in Section II are provided in Appendix A.)}
\end{itemize}

For accurately quantifying the error in recovering the underlying binary matrix, we propose to adopt the $\omega$-weighted square loss \cite{b37,b38} as the loss function. The $\omega$-weighted square loss is defined as
\begin{equation}
{\ell _\omega }\left( {x,a} \right) = \omega {I_{a = {a_1}}}\ell \left( {x,{a_1}} \right) + \left( {1 - \omega } \right){I_{a = {a_2}}}\ell \left( {x,{a_2}} \right).\label{eq3}
\end{equation}
where $\ell \left( {x,a} \right) = {\left( {x - a} \right)^2}$ is the square loss, ${I_{a = {a_1}}}$ and ${I_{a = {a_2}}}$ are indicator functions, i.e., ${I_{a = {a_1}}}$ is 1 if $a = {a_1}$ is true and 0 otherwise.

Consequently, \eqref{eq2} can be further formulated as follows.
\begin{equation}
\mathop {\min }\limits_{{\mathbf{X}},{\mathbf{A}} \in {{\left\{ {0,1} \right\}}^{m \times n}}} F\left( {\mathbf{X}} \right) \equiv \lambda {r_n}\left( {\mathbf{X}} \right) + \sum\nolimits_{i,j} {{\ell _\omega }\left( {{X_{ij}},{A_{ij}}} \right)}.\label{eq4}
\end{equation}
where ${\mathbf{X}}$ and ${\mathbf{A}}$ are the recovery matrix and observation matrix of the underlying target matrix $\mathbf{M}$, respectively, and ${\ell _\omega }\left( {{X_{ij}},{A_{ij}}} \right) = \omega {I_{{A_{ij}} = 1}}\ell \left( {{X_{ij}},1} \right) + \left( {1 - \omega } \right){I_{{A_{ij}} = 0}}\ell \left( {{X_{ij}},0} \right)$.

\textbf{Recovery error of \eqref{eq4}.} Following the definition of the recovery error in \cite{b16,b18}, here the recovery error can be formulated as
\begin{equation}
R\left( {\mathbf{X}} \right) = \frac{1}{{mn}}\left\| {{\mathbf{X}} - {\mathbf{M}}} \right\|_F^2.\label{eq5}
\end{equation}
where ${\mathbf{M}},{\mathbf{X}} \in {\mathbb {R}^{m \times n}}$ is the underlying target matrix and its recovery matrix, respectively.

In \cite{b38}, the label-dependent loss is defined as $U\left( {x,a} \right) = {I_{x = 1}}{I_{a = 0}} + {I_{x = 0}}{I_{a = 1}}$ . And similar to \eqref{eq3}, we define the weighted version of the label-dependent loss as 
\begin{equation}
{U_\omega }\left( {x,a} \right) = \left( {1 - \omega } \right){I_{x = 1}}{I_{a = 0}} + \omega {I_{x = 0}}{I_{a = 1}}.\label{eq6}
\end{equation}

Therefore, the corresponding $\omega$-weighted expected error can be written as
\begin{equation}
{R_\omega }\left( {\mathbf{X}} \right) = E\left[ {\sum\nolimits_{i,j} {{U_\omega }\left( {{X_{ij}},{A_{ij}}} \right)} } \right].\label{eq7}
\end{equation}

According to the class-conditional random noise model in \cite{b39}, \eqref{eq} can be written correspondingly as
\begin{equation}
\begin{array}{l}
P\left( {\left. {{A_{ij}} = 0} \right|{M_{ij}} = 1} \right) = 1 - \delta  = {\rho _{ + 1}}\\
P\left( {\left. {{A_{ij}} = 1} \right|{M_{ij}} = 0} \right) = 0 = {\rho _{ - 1}}
\end{array}.\label{eq8}
\end{equation}
\\
\textbf{Theorem 1}. \textit{For the choices $\hat \omega  = \frac{{1 - {\rho _{ + 1}}}}{2}$ and $\kappa  = \frac{{1 - {\rho _{ + 1}}}}{2}$, there exists a constant $c$ that is dependent of ${\mathbf{X}}$ such that, for any matrix ${\mathbf{X}}$, we have ${R_{\hat \omega }}\left( {\mathbf{X}} \right) = \kappa R\left( {\mathbf{X}} \right) + c$.}
\\

\begin{table}[t]
	\centering
	\caption{The functions $r$ and thresholds $\gamma$  for nonconvex regularizers where $\mu > 0$, $\eta  = {\textstyle{\lambda  \over \rho }}$. For the TNN regularizer, $\mu$ is an integer denoting the number of leading singular values that are not penalized.}
	\label{table1}
	\setlength{\tabcolsep}{2mm}{
		\begin{tabular}{lll} 
			\toprule 
			& $\eta r\left( {{\sigma _i}\left( {\mathbf{X}} \right)} \right)$  
			& \multicolumn{1}{c}{$\gamma$}\\
			\midrule 
			TNN    
			& $\left\{ \begin{array}{c}
			0   {\kern 1pt} {\kern 1pt} {\kern 1pt} {\kern 1pt} {\kern 1pt} {\kern 1pt} {\kern 1pt} {\kern 1pt} {\kern 1pt} {\kern 1pt} {\kern 1pt} {\kern 1pt} {\kern 1pt} {\kern 1pt} {\kern 1pt} {\kern 1pt} {\kern 1pt} {\kern 1pt} {\kern 1pt} {\kern 1pt} {\kern 1pt} {\kern 1pt} {\kern 1pt} {\kern 1pt} {\kern 1pt} {\kern 1pt} ,i \le \mu \\
			\eta {\sigma _i}\left( {\mathbf{X}} \right),i > \mu 
			\end{array} \right.$ 
			&$\max \left( {{\sigma _{\mu  + 1}}\left( {\mathbf{X}} \right),\eta } \right)$  \\[0.3cm]
			capped $\ell_1$
			& $\left\{ \begin{array}{c}
			\eta {\sigma _i}\left( {\mathbf{X}} \right),{\sigma _i}\left( {\mathbf{X}} \right) \le \mu \\
			\eta \mu {\kern 1pt} {\kern 1pt} {\kern 1pt} {\kern 1pt} {\kern 1pt} {\kern 1pt} {\kern 1pt} {\kern 1pt} {\kern 1pt} {\kern 1pt} {\kern 1pt} {\kern 1pt} {\kern 1pt} {\kern 1pt} {\kern 1pt} {\kern 1pt} {\kern 1pt} {\kern 1pt} {\kern 1pt} {\kern 1pt} ,{\sigma _i}\left( {\mathbf{X}} \right) > \mu 
			\end{array} \right.$ 
			&$\min \left( {\eta ,\sqrt {2\mu \eta } } \right)$  \\[0.3cm]
			LSP    
			&$\eta \log \left( {{{{\sigma _i}\left( {\mathbf{X}} \right)} \mathord{\left/
						{\vphantom {{{\sigma _i}\left( {\mathbf{X}} \right)} \mu }} \right.
						\kern-\nulldelimiterspace} \mu } + 1} \right)$  
			& $\min \left( {{\eta  \mathord{\left/
						{\vphantom {\eta  \mu }} \right.
						\kern-\nulldelimiterspace} \mu },\mu } \right)$ \\
			\bottomrule 
	\end{tabular}}
\end{table}
The above theorem (Theorem 1) is a special case of Theorem 9 in \cite{b39}. At this juncture, the linear mapping between the recovery error $R\left( {\mathbf{X}} \right)$ and $\omega$-weighted expected error ${R_{\hat \omega }}\left( {\mathbf{X}} \right)$ indicates that minimizing $R\left( {\mathbf{X}} \right)$ is equivalent to minimizing ${R_{\hat \omega }}\left( {\mathbf{X}} \right)$ on the partial information.\\
\\
\textbf{Theorem 2} (Main Result 1). \textit{Let ${\mathbf{\hat X}} \in {\mathbb{R}^{m \times n}}$ be the solution to \eqref{eq4}, then with probability at least $1-\alpha$,}
\begin{equation}
\frac{1}{{mn}}\left\| {{\mathbf{\hat X}} - {\mathbf{M}}} \right\|_F^2 \le \frac{C}{{{\delta ^2}}}\left( {\sqrt {\frac{{\log \left( {{2 \mathord{\left/
							{\vphantom {2 \alpha }} \right.
							\kern-\nulldelimiterspace} \alpha }} \right)}}{{mn}}}  + b} \right).\label{eq9}
\end{equation}
\textit{where $C$ is absolute constant, $\delta$ denotes the sampling rate, $b = \frac{{\sqrt m  + \sqrt n  + \sqrt[4]{{\left| {{\Omega _1}} \right|}}}}{{mn}}$, $\Omega_1$ is the index set of all positive elements in ${\mathbf{M}}$}. The proof can be found in Appendix C.

\section{Algorithm}
In this section, we will show the nonconvex model can be solved much faster. First, Subsection A shows the basic algorithm for nonconvex regularizers, followed by acceleration measures in Subsection B. Subsection C summarizes the whole algorithm and Subsection D analyses the resultant algorithm from the aspect of convergence and time complexity.
\subsection{Basic algorithm}
Let ${L_\omega }\left( {{\mathbf{X}},{\mathbf{A}}} \right) = \sum\nolimits_{i,j} {{\ell _\omega }\left( {{X_{ij}},{A_{ij}}} \right)}$, and in line with the definition of the Frobenius norm, we have the following derivation.
\begin{equation}
\begin{array}{l}
{L_\omega }\left( {\mathbf{X},\mathbf{A}} \right)\\=
\left( {1 - \omega } \right)\sum\limits_{{A_{ij}} = 0} {{{\left( {{X_{ij}} - {A_{ij}}} \right)}^2}}  + \omega \sum\limits_{{A_{ij}} = 1} {{{\left( {{X_{ij}} - {A_{ij}}} \right)}^2}} \\
= \left( {1 - \omega } \right)\left\| {\mathbf{X} - \mathbf{A}} \right\|_F^2 + \left( {2\omega  - 1} \right)\left\| {{{\cal P}_\Omega }\left( {\mathbf{X} - \mathbf{A}} \right)} \right\|_F^2
\end{array}.\label{eq10}
\end{equation}
where ${\left[ {{{\cal P}_\Omega }\left( {{\mathbf{X}} - {\mathbf{A}}} \right)} \right]_{ij}} = {\left( {{\mathbf{X}} - {\mathbf{A}}} \right)_{ij}}$ if $\left( {i,j} \right) \in \Omega $ and 0 otherwise. In recent years, the proximal algorithm \cite{b32} has been regarded as an efficient method for solving the optimization problem $\mathop {\min }\limits_\mathbf{X} {f_1}\left( \mathbf{X} \right) + {f_2}\left( \mathbf{X} \right)$, when $f_1$ and $f_2$ are convex. The following theorem shows that the convergence of the proximal algorithm.\\
\\
\textbf{Theorem 3} \cite{b40}. \textit{Let $f_1$, $f_2$ be lower semicontinuous, and $f_1$ is differentiable with $\beta$-Lipschitz continuous gradient. If $f_1 + f_2$ is coercive and strictly convex, the solution of the problem takes on uniqueness. For an arbitrary initial matrix $\mathbf{X}$, $\forall \rho  \ge \beta $, the iterative sequence   generated by the following statement can converge to the unique solution of the problem.}
\begin{equation}
{{\mathbf{X}}_{t + 1}} = pro{x_{{\textstyle{\lambda \over \rho }}{f_2}}}\left( {{{\mathbf{X}}_t} - {\textstyle{1 \over \rho }}\nabla {f_1}\left( {{{\mathbf{X}}_t}} \right)} \right).\label{eq11}
\end{equation}
\textit{where $pro{x_{{\textstyle{\lambda \over \rho }}{f_2}}}\left( {\mathbf{Z}} \right) = \mathop {\arg \min }\limits_{{\mathbf{X}},{\mathbf{Z}} \in {\mathbb{R}^{m \times n}}} \left\{ {\frac{1}{2}\left\| {{\mathbf{X}} - {\mathbf{Z}}} \right\|_F^2 + {\textstyle{\lambda \over \rho }}{f_2}\left( {\mathbf{X}} \right)} \right\}$ denotes the proximal operator.}
\\

If $f_2$ is the nuclear norm, the following theorem shows that the proximal operator of the nuclear norm has a closed form solution.\\
\\
\textbf{Theorem 4} \cite{b33}. \textit{For an arbitrary matrix ${\mathbf{Z}} \in {\mathbb{R}^{m \times n}}$, $\forall \tau  > 0$, the proximal operator of the nuclear norm of matrix ${\mathbf{X}}$ is}
\begin{equation}
pro{x_{\tau {{\left\| {\mathbf{X}} \right\|}_*}}}\left( {\mathbf{Z}} \right) = {\mathbf{U}}{\left( {{\mathbf{\Sigma }} - \tau {\mathbf{I}}} \right)_ + }{{\mathbf{V}}^\top }.\label{eq12}
\end{equation}
\textit{where $\mathbf{I}$ denotes the identity matrix, $SVD\left( {\mathbf{Z}} \right) = {\mathbf{U\Sigma }}{{\mathbf{V}}^\top }$, and ${\left[ {{\mathbf{M}_ + }} \right]_{ij}} = max\left( {{M_{ij}},0} \right)$.}
\\

For solving \eqref{eq4}, we extend the proximal operator to nonconvex problem, similar to Theorem 3, at $t$-iteration, it products the iterative sequence as follows.
\begin{equation}
{{\mathbf{X}}_{t + 1}} = pro{x_{{\textstyle{\lambda  \over \rho }}{r_n}}}\left( {{{\mathbf{X}}_t} - {\textstyle{1  \over \rho }}\nabla {L_\omega }\left( {{{\mathbf{X}}_t},{\mathbf{A}}} \right)} \right).\label{eq13}
\end{equation}
where the learning rate, herein, denoted by $\rho$ is a fixed value, and $\nabla {L_\omega }\left( {{\mathbf{X}},{\mathbf{A}}} \right)$ denotes the gradient of the $\omega$-weighted loss function, which can be computed efficiently as
\begin{equation}
{\textstyle{1 \over 2}}\nabla {L_\omega }\left( {{\mathbf{X}},{\mathbf{A}}} \right) = \left( {1 - \omega } \right)\left( {{\mathbf{X}} - {\mathbf{A}}} \right) + \left( {2\omega  - 1} \right){{\cal P}_\Omega }\left( {{\mathbf{X}} - {\mathbf{A}}} \right).\label{eq14}
\end{equation}

Recently, due to the successful application on convex optimization problem, the proximal algorithm has been extended to nonconvex situation \cite{b9,b13,b14,b20}. Similar to the nuclear norm (Theorem 4), the generalized singular value thresholding \cite{b20} was proposed to handle the nonconvex surrogates.\\
\\
\textbf{Theorem 5} Generalized singular value thresholding (GSVT) \cite{b20}. \textit{For an arbitrary matrix ${\mathbf{Z}} \in {\mathbb{R}^{m \times n}}$, let $r_n$ be a function that satisfies the characteristics in \eqref{eq2}, then the proximal operator of $r_n$ has the following closed form solution.}
\begin{equation}
pro{x_{{r_n}}}\left( {\mathbf{Z}} \right) = {\mathbf{U}}diag\left( {\hat s} \right){{\mathbf{V}}^\top}.\label{eq15}
\end{equation}
\textit{where ${\mathbf{U}}{\mathbf{\Sigma}}{{\mathbf{V}}^\top}$ is the SVD of $\mathbf{Z}$, and $\hat s = \left\{ {{{\hat s}_i}} \right\}$ with}
\begin{equation}
{\hat s_i} \in \arg \mathop {\min }\limits_{{s_i} \ge 0} {\textstyle{1 \over 2}}{\left( {{s_i} - {\sigma _i}\left( {\mathbf{Z}} \right)} \right)^2} + {\textstyle{\lambda  \over \rho }}r\left( {{s_i}} \right).\label{eq16}
\end{equation}

Similar to Theorem 4, the above theorem indicates that the closed-form solutions of the nonconvex regularizer in Table \ref{table1} do exist. In addition, we generalize the above procedure as the Basic Algorithm shown in Algorithm 1.

\begin{algorithm}[ht]
	\caption{Basic Algorithm}
	\LinesNumbered
	\KwIn{${\mathbf{A}} \in {\mathbb{R}^{m \times n}}$, $\rho > \beta$, the sampling set $\Omega$, and the regularization parameter $\lambda$.}
	initialize ${{\mathbf{X}}_1} = 0$;\\
	\For{${\rm{t}} = 1,2,...,T$}{
		${{\mathbf{X}}_{ig}} = {{\mathbf{X}}_t} - {\textstyle{1 \over \rho }}\nabla {L_\omega }\left( {{{\mathbf{X}}_t},{\mathbf{A}}} \right);$\
		
		$[{\mathbf{U}},{\mathbf{\Sigma }},{\mathbf{V}}] = SVD\left( {{{\mathbf{X}}_{ig}}} \right);$
		
		\For{${\rm{i}} = 1,2,...,m$}{
			${\hat s_i} \in \arg \mathop {\min }\limits_{{s_i} \ge 0} {\textstyle{1 \over 2}}{\left( {{s_i} - {\Sigma _{ii}}} \right)^2} + {\textstyle{\lambda  \over \rho }}r\left( {{s_i}} \right);$
		}
	$\mathbf{X}_t = \mathbf{U} diag\left(\left\{\hat{s}_i\right\}\right) \mathbf{V}^\top$;
	}
	\KwResult{${{\mathbf{X}}_{T + 1}}$}
\end{algorithm}

\subsection{Acceleration}
However, the basic algorithm involves a full SVD (step 3) with the time complexity ${{\cal O}}\left( {m{n^2}} \right)$. Next, we will take the following two schemes to make the basic algorithm much faster.

\textbf{S1}. The first thing that comes to our mind is to use partial SVD. However, $\hat s_i$ in \eqref{eq16}  actually becomes zero when the singular value ${\sigma _i}\left( {\mathbf{Z}} \right)$ is not larger than a threshold $\gamma$, that is automatic thresholding in \cite{b41}. This means only the leading few singular values, instead of all singular values, are needed to compute the proximal operator in Theorem 5. The thresholds $\gamma$ for the mentioned nonconvex regularizers are shown in Table \ref{table1}.

\textbf{S2}. Another measure our carry out is reducing the size of SVD. The following theorem shows that the proximal operator on a large matrix can be replaced by the counterpart on a smaller one. The proof can be found in Appendix D.
\\
\\
\textbf{Theorem 6}. \textit{For an arbitrary matrix $\mathbf{Z} \in {\mathbb{R}^{m \times n}}$, let $k$ be the number of singular values of $\mathbf{Z}$ that are not less than $\gamma$, its rank-$k$ SVD is ${{\mathbf{U}}_k}{{\mathbf{\Sigma }}_k}{\mathbf{V}}_k^\top$, and $\mathbf{W} \in {\mathbb{R}^{m \times k}}$ be an orthogonal matrix, if $span\left( {{{\mathbf{U}}_k}} \right) \subseteq span\left( {\mathbf{W}} \right)$ ,then the following equation holds.}
\begin{equation}
pro{x_{{r_n}}}\left( {\mathbf{Z}} \right) = {\mathbf{W}} pro{x_{{r_n}}}\left( {{\mathbf{W}}{ ^{ \top }}{\mathbf{Z}}} \right).\label{eq17}
\end{equation}

Theorem 6 performs the process of replacing a large matrix by its projection on leading subspace. How to obtain such $\mathbf{W}$ in Theorem 6? Two approaches are available to find the $\mathbf{W}$ exactly in the same time complexity. The first method, PROPACK package \cite{b42}, is widely applied to partial SVD. And the second one is the power method which has good approximation guarantee \cite{b43}. Compared with the former method, the latter one can benefit  particularly from warm-start taking full advantage of the iterative nature of the proximal algorithm. Hence, we use the power method to get the $\mathbf{W}$, and the details are shown in Algorithm 2.

\begin{algorithm}
	\caption{power method}
	\LinesNumbered
	\KwIn{ Let ${\mathbf{Z}} \in {\mathbb{R}^{m \times n}}$, ${\mathbf{Y}} \in {\mathbb{R}^{n \times k}}$, and the number of iterations $H$.}
	${{\mathbf{R}}_1} = {\mathbf{ZY}};$
	
	\For{${\rm{h}} = 1,2,...,H$}{
		${{\mathbf{W}}_h} = QR\left( {{{\mathbf{R}}_h}} \right);$//QR decomposition\\
		${{\mathbf{R}}_{h + 1}} = {\mathbf{Z}}\left( {{{\mathbf{Z}}^{ \top }}{{\mathbf{W}}_h}} \right);$
	}
	\KwResult{${{\mathbf{W}}_{H}}$}
\end{algorithm}

Let ${{\mathbf{X}}_{ig}} = {{{\mathbf{X}}_t} - {\textstyle{1  \over \rho }}\nabla {L_\omega }\left( {{{\mathbf{X}}_t},{\mathbf{A}}} \right)} $, through the implementation of the above two acceleration measures, \eqref{eq13} can be rewritten as
\begin{equation}
{{\mathbf{X}}_{t + 1}} = {\mathbf{W}}{{\mathbf{U}}_a}diag(\hat s){\mathbf{V}}_a^{ \top }.\label{eq18}
\end{equation}
where ${{\mathbf{U}}_a}{{\mathbf{\Sigma }}_a}{\mathbf{V}}_a^\top$ is the rank-$a$ SVD of ${\mathbf{W}^{\top} {\mathbf{X}}_{ig}}$,$a$ is the number of singular values that are greater than the threshold $\gamma$, and $\hat s$ can be obtained from \eqref{eq16}.

\subsection{The whole algorithm}
We summarize the whole procedure for solving \eqref{eq4} in Algorithm 3 and name it PUMC\_N (\underline{P}ositive and \underline{U}nlabeled \underline{M}atrix \underline{C}ompletion with \underline{N}onconvex regularizers). In step 3, similar to the nmAPG algorithm \cite{b44}, a linear combination of ${{\mathbf{X}}_{t - 1}}$ and ${{\mathbf{X}}_t}$ is used to accelerate the algorithm. The column spaces of the current iteration (${{\mathbf{V}}_t}$) and previous iteration (${{\mathbf{V}}_{t - 1}}$) are used to accomplish the warm start in step 5 as in \cite{b32}. Step 6 and step 7 performs \textbf{S2}, and in step 8, a continuation strategy is introduced to speed up  the algorithm further. Specifically, as the iteration proceeds, $\lambda$ is dynamic and gradually decreases from a large value. In addition, step 9-11 performs \textbf{S1}.

\begin{algorithm}
	\caption{PUMC\_N}
	\LinesNumbered
	\KwIn{ Let ${\lambda _0} > \lambda$, $\rho > \beta$, $\upsilon $, $\mathbf{A} \in \mathbb{R} ^ {m \times n}$, and the sampling set $\Omega$.}
     Initialize ${{\mathbf{X}}_0} = {{\mathbf{X}}_1} = 0$, $\alpha_0 = \alpha_1 = 1$, and ${{\mathbf{V}}_{0}},{{\mathbf{V}}_1} \in {\mathbb{R}^{n \times 1}}$ as random Gaussian matrices;
	
	\For{${\rm{t}} = 1,2,...,T$}{
		${{\mathbf{Z}}_t} = {{\mathbf{X}}_t} + \frac{{{\alpha _{t - 1}} - 1}}{{{\alpha _t}}}\left( {{{\mathbf{X}}_t} - {{\mathbf{X}}_{t - 1}}} \right);$
		
		${{\mathbf{Z}}_{ig}} = {{\mathbf{Z}}_t} - {\textstyle{1 \over \rho }}\nabla {L_\omega }\left( {{{\mathbf{Z}}_t},{\mathbf{A}}} \right);$
		
		${{\mathbf{Y}}_t} = QR\left( {\left[ {{{\mathbf{V}}_t},{{\mathbf{V}}_{t - 1}}} \right]} \right);$
		
		${\mathbf{W}} = powermethod\left( {{{\mathbf{Z}}_{ig}},{{\mathbf{Y}}_t}} \right);$
		
		$\left[ {{\mathbf{U}},{\mathbf{\Sigma }},{\mathbf{V}}} \right] = SVD\left( {{{\mathbf{W}}^ \top }{{\mathbf{Z}}_{ig}}} \right);$
		
		${\lambda _t} = \left( {{\lambda _{t - 1}} - \lambda } \right){\upsilon ^t} + \lambda ;$
		
		\For{${\rm{i}} = 1,2,...,k$}{
			${\hat s_i} \in \arg \mathop {\min }\limits_{{s_i} \ge 0} {\textstyle{1 \over 2}}{\left( {{s_i} - {\Sigma _{ii}}} \right)^2} + {\textstyle{\lambda_t  \over \rho }}r\left( {{s_i}} \right);$
		}
	${{\mathbf{X}}_{t + 1}} = {\mathbf{W}}{{\mathbf{U}}_k}diag\left( {\left\{{\hat s}_i \right\}} \right){\mathbf{V}}_k^ \top;$
	
	${{\mathbf{V}}_{t + 1}} = {\mathbf{V}};$
	
	${\alpha _{t + 1}} = {\textstyle{1 \over 2}}\left( {\sqrt {4\alpha _t^2 + 1}  + 1} \right);$
	}
	\KwResult{${{\mathbf{X}}_{T + 1}}$}
\end{algorithm}

\subsection{Algorithm analysis}
\textbf{Convergence analysis}. Firstly, we present a lemma which provides the basic support for the convergence analysis of the proposed algorithm. The following lemma shows that the objective function $F$ is nonincreasing as iterations proceed.
\\
\\
\textbf{Lemma 1} \cite{b45}. \textit{Let $\left\{ {{{\mathbf{X}}_t}} \right\}$ be the iterative sequence produced by \eqref{eq13}, for the optimization problem \eqref{eq2}, we have $F\left( {{{\mathbf{X}}_{t + 1}}} \right) \!\le\! F\left( {{{\mathbf{X}}_t}} \right) \!-\! {\textstyle{{\rho - \beta } \over 2}}\left\| {{{\mathbf{X}}_{t + 1}} \!-\! {{\mathbf{X}}_t}} \right\|_F^2$, where $\rho \!>\! \beta$.}
\\

The following theorem shows that the proposed algorithm generates a bounded iterative sequence. The proof can be found in Appendix E.
\\
\\
\textbf{Theorem 7}. \textit{Let $\left\{ {{{\mathbf{X}}_t}} \right\}$ be the iterative sequence produced by \eqref{eq13}, we say $\left\{ {{{\mathbf{X}}_t}} \right\}$ is a bounded iterative sequence, i.e., $\sum\nolimits_{t = 1}^\infty  {\left\| {{{\mathbf{X}}_{t + 1}} - {{\mathbf{X}}_t}} \right\|_F^2}  < \infty $.}
\\

For the convex optimization problem in Theorem 3, the proximal mapping in  \cite{b45} is denoted by ${G_{{\scriptstyle{1 \over \rho }}{f_2}}}\left( {{{\mathbf{X}}_t}} \right) = pro{x_{{\scriptstyle{1 \over \rho }}{f_2}}}\left( {{{\mathbf{X}}_t} - {\textstyle{1 \over \rho }}\nabla {f_1}\left( {{{\mathbf{X}}_t}} \right)} \right) - {{\mathbf{X}}_t}$. In particular, when $f_2$ is convex, ${\left\| {{G_{{\scriptstyle{1 \over \rho }}{f_2}}}\left( {{{\bf{X}}_t}} \right)} \right\|_2^2}$ can be used to conduct the convergence analysis. In contrast, if $f_2$ is nonconvex, it is no longer applicable. Hence, we use $\left\| {{G_{{\scriptstyle{1 \over \rho }}{f_2}}}\left( {{{\mathbf{X}}_t}} \right)} \right\|_F^2{\rm{ = }}\left\| {{{\mathbf{X}}_{t + 1}} - {{\mathbf{X}}_t}} \right\|_F^2$ instead to perform convergence analysis of the proposed algorithm. The convergence of Algorithm 3 is shown in the following theorem, and the proof can be found in Appendix F.
\\
\\
\textbf{Theorem 8} (Main Result 2). \textit{Let $\left\{ {{{\mathbf{X}}_t}} \right\}$ be the iterative sequence in Algorithm 3, for the consecutive elements ${\mathbf{X}_{t}}$ and ${\mathbf{X}_{t + 1}}$, we have}

\begin{equation}
\mathop {\min }\limits_{t = 1, \cdots ,T} \left\| {{{\mathbf{X}}_{t + 1}} - {{\mathbf{X}}_t}} \right\|_F^2 \le {\textstyle{2 \over {\left( {\rho  - \beta } \right)T}}}\left( {F\left( {{{\mathbf{X}}_1}} \right) - \inf F} \right).\label{eq19}
\end{equation}
\\
\textbf{Time complexity}. Assume that $\mathbf{Y}_{t}$ in step 5 of Algorithm 3 has $k_t$ columns at the current iteration. Consequently, step 5 takes ${{\cal O}}\left( {nk_t^2} \right)$ time. Next, step 3 shows that $\mathbf{Z}_{t}$ is a linear combination of $\mathbf{X}_{t - 1}$ and $\mathbf{X}_{t}$. Let ${c_t} = {{\left( {{\alpha _{t - 1}} - 1} \right)} \mathord{\left/
		{\vphantom {{\left( {{\alpha _{t - 1}} - 1} \right)} {{\alpha _t}}}} \right.
		\kern-\nulldelimiterspace} {{\alpha _t}}}$, combining step 4 and step 5, we have 
\begin{equation}
{{\mathbf{Z}}_{ig}} = \left\{ {{c_1}{{\mathbf{X}}_t} + {c_2}{{\mathbf{X}}_{t - 1}} + {c_3}{\mathbf{A}}} \right\} + {c_4}{{\cal P}_\Omega }\left( {{{\mathbf{Z}}_t} - {\mathbf{A}}} \right) .\label{eq20}
\end{equation}
where ${c_1} = \left( {1 + {c_t}} \right)\left( {1 - {c_3}} \right)$, ${c_2} = {c_t}\left( {{c_3} - 1} \right)$, ${c_3} = {\textstyle{{2\left( {1 - \omega } \right)} \over \rho }}$, ${c_4} = {\textstyle{{2\left( {1 - 2\omega } \right)} \over \rho }}$. The first three terms involve low-rank matrices, whereas the last term involves a sparsity structure. The combined structure in \eqref{eq20} was studied specifically in \cite{b42}. Consider the multiplication of $\mathbf{Z}_{ig}$ and a vector $\mathbf{b} \in {\mathbb{R}^{n}}$. For the low-rank part the multiplication cost ${{\cal O}}\left( {\left( {m + n} \right){k_t}} \right)$ time, whereas the sparse part cost ${{\cal O}}\left( {{{\left\| \Omega  \right\|}_1}} \right)$ time. Hence, the cost is ${{\cal O}}\left( {\left( {m + n} \right){k_t} + {{\left\| \Omega  \right\|}_1}} \right)$ per vector multiplication. Step 8 performs a rank-$k_t$ SVD of ${{\mathbf{W}}^ \top }{{\mathbf{Z}}_{ig}}$, it takes  ${{\cal O}}\left( {\left( {m + n} \right)k_t^2 + {{\left\| \Omega  \right\|}_1}{k_t}} \right)$ time. In summary, the order of the time complexity at the current iteration is ${{\cal O}}\left( {\left( {m + n} \right)k_t^2 + {{\left\| \Omega  \right\|}_1}{k_t}} \right)$.

\section{Experiments}
In this section, we perform experiments on synthetic and real-world data sets and demonstrate the effectiveness of the proposed algorithm in practical applications, including link prediction, topology inference, and recommender system. All experiments are implemented in Matlab on Windows 10 with Intel Xeon CPU (2.8GHz) and 128GB memory.

\subsection{Synthetic Data}
\textbf{Data sets}. As in \cite{b18,b41}, we assume the matrix $\mathbf{Q} \in {\mathbb{R} ^{m \times m}}$ is generated by $\mathbf{Q} = \mathbf{M}_1 \mathbf{M}_2$, where the elements of $\mathbf{M}_1 \in {\mathbb{R} ^{m \times k}}$ and $\mathbf{M}_2 \in {\mathbb{R} ^{k \times m}}$ are obtained from the Gaussian distribution ${\cal N}\left( {0,1} \right)$. Therefore, the underlying binary matrix $\mathbf{M} \in {\mathbb{R} ^{m \times m}}$ can be generated by ${M_{ij}} = {{I}_{{Q_{ij}} \ge q}}$, without loss of generality, we assume that $q = 0.5$. We fix $k = 5$ and very $m$ in $\left\{ {50, 100, 500, 1000, 2000} \right\}$.

For each scenario, the mean square error ${\rm{MSE}} = {{\left\| {{{\cal P}_{\bar \Omega }}\left( {{\mathbf{X}} - {\mathbf{M}}} \right)} \right\|_F^2} \mathord{\left/
		{\vphantom {{\left\| {{{\cal P}_{\bar \Omega }}\left( {{\bf{X}} - {\mathbf{M}}} \right)} \right\|_F^2} {\left\| {{{\cal P}_{\bar \Omega }}\left( {\mathbf{M}} \right)} \right\|_F^2}}} \right.
		\kern-\nulldelimiterspace} {\left\| {{{\cal P}_{\bar \Omega }}\left( {\mathbf{M}} \right)} \right\|_F^2}}$ is used for performance evaluation, where $\mathbf{X}$ is the recovery matrix for underlying matrix $\mathbf{M}$ and  $\bar \Omega $ is the index set of unobserved elements. Each experiment is repeated ten times with the sampling rate ($\delta $) varying from 0.3 to 0.9, and the average results are reported. From Theorem 1, if the sampling rate $\delta = 0.3 $ (only 30\% 1's in $\mathbf{M}$ are observed), $\omega = 0.15$   is chosen.

\begin{figure}[ht]
	\centering
	\subfigure[]{
		\includegraphics[width=0.45\textwidth]{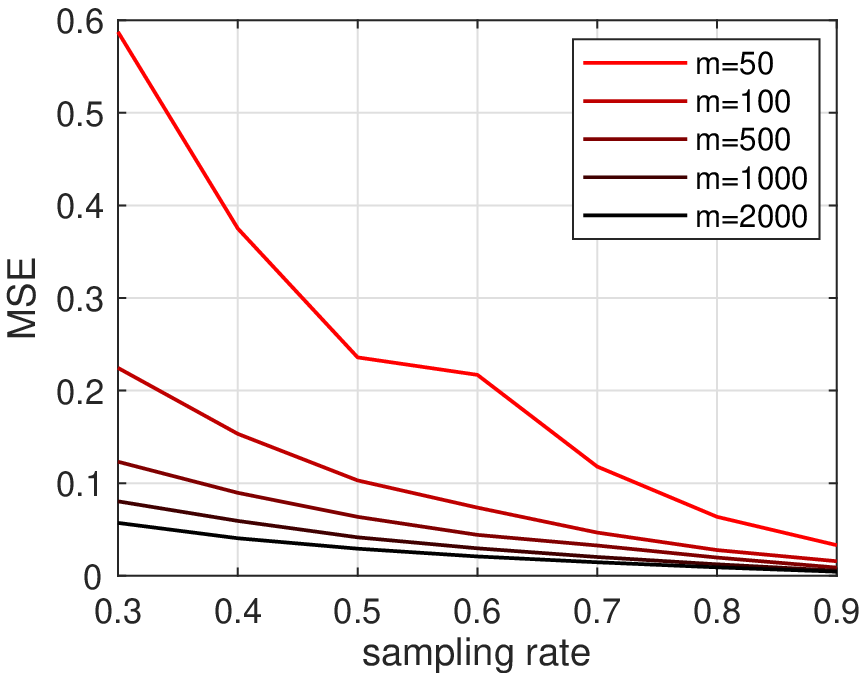}
		\label{subfig1}}
	\subfigure[]{
		\includegraphics[width=0.45\textwidth]{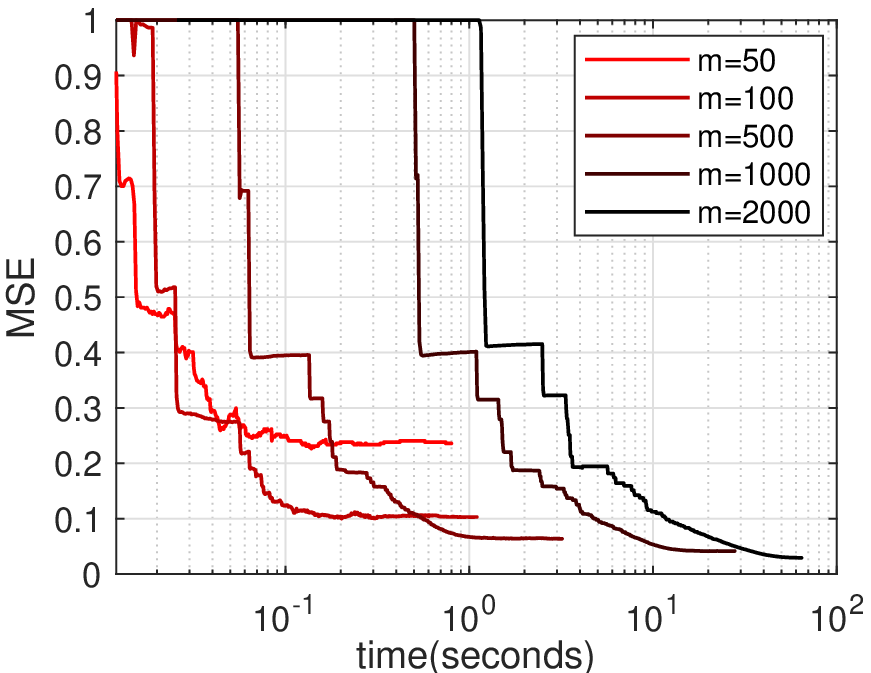}
		\label{subfig2}}
	\caption{Performance analysis of the proposed PUMC\_N algorithm on synthetic datasets. (a) MSE vs sampling rate on synthetic datasets. (b) The sampling rate is fixed at 0.5, MSE vs time (in seconds) on synthetic data sets.}
	\label{Fig.1}
\end{figure}

Results are shown in Fig. \ref{Fig.1}. Only TNN regularizer (with $\mu $ in Table \ref{table1} set to 5) is used in synthetic experiments, similar results can be obtained from other nonconvex regularizers in Table \ref{table1}. Fig. \ref{subfig1} shows the testing MSE at different sampling rates. Notice that for an increasing sampling rate we see a monotonous decrease in testing MSE until it is close to zero. This is reasonable since a larger number of observations give rise to more accurate information of the underlying matrix. In addition, there is also a negative correlation between MSE and the matrix size, which is particularly evident at a smaller sampling rate. In particular, if the underlying matrix $\mathbf{M}$ is large enough, it can be recovered accurately at a much small sampling rate (with small number of observations). In Fig. \ref{subfig2}, we follow the settings in \cite{b41} and fix the sampling rate at 0.5 (from Theorem 1, $\omega  = 0.25$). It can be seen that the MSE drops sharply and precipitously at the beginning. Moreover, the larger the matrix, the more time the algorithm takes, and the smaller the MSE.

\subsection{Recommender System}
\textbf{Data sets}. In this practical setting, the popular data sets (Table \ref{table5}), including \textit{MovieLens (100K)}\footnote{http://vladowiki.fmf.uni-lj.si/doku.phpid=pajek:data:pajek:students} \cite{b16}, \textit{ FlimTrust}\footnote{https://www.librec.net/datasets.html} \cite{b53}, and \textit{Douban}\footnote{https://github.com/fmonti/mgcnn} \cite{b54}, are used to evaluate the performance of our algorithm. We follow the setup in \cite{b16,b19} and convert these ratings in each data set to binary observations by comparing each rating to the average value (which is $\sim 3$, considering three data sets together) of whole data sets.
\\
\textbf{Methods}. We compare with the nuclear norm based algorithm AIS-Impute \cite{b32}, as well as 1-bit matrix completion in \cite{b19}. In particular, the AIS-Impute can be considered as an accelerated and inexact version of the proximal algorithm, and the 1-bit matrix completion is constrained by infinity norm and nuclear norm.
\begin{figure*}
	\centering
	\subfigure[]{
		\includegraphics[width=0.45\textwidth]{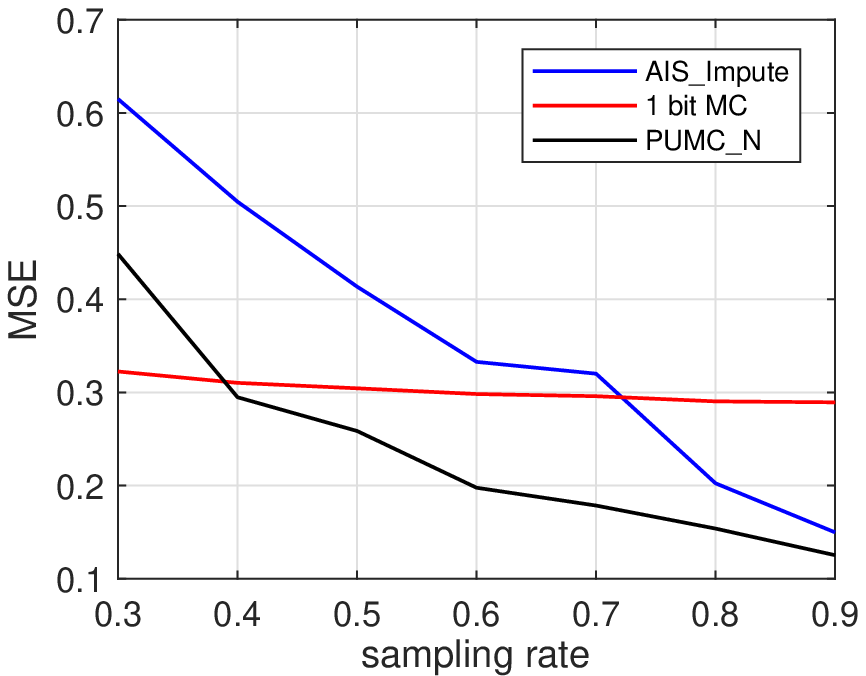}
		\label{subfig9}}
	\subfigure[]{
		\includegraphics[width=0.45\textwidth]{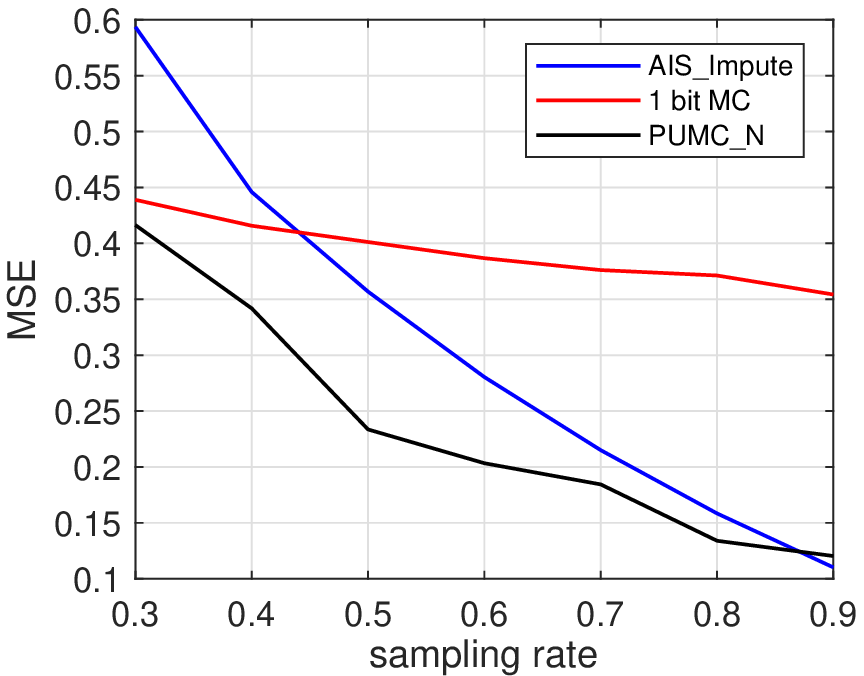}
		\label{subfig10}}
	\subfigure[]{
		\includegraphics[width=0.45\textwidth]{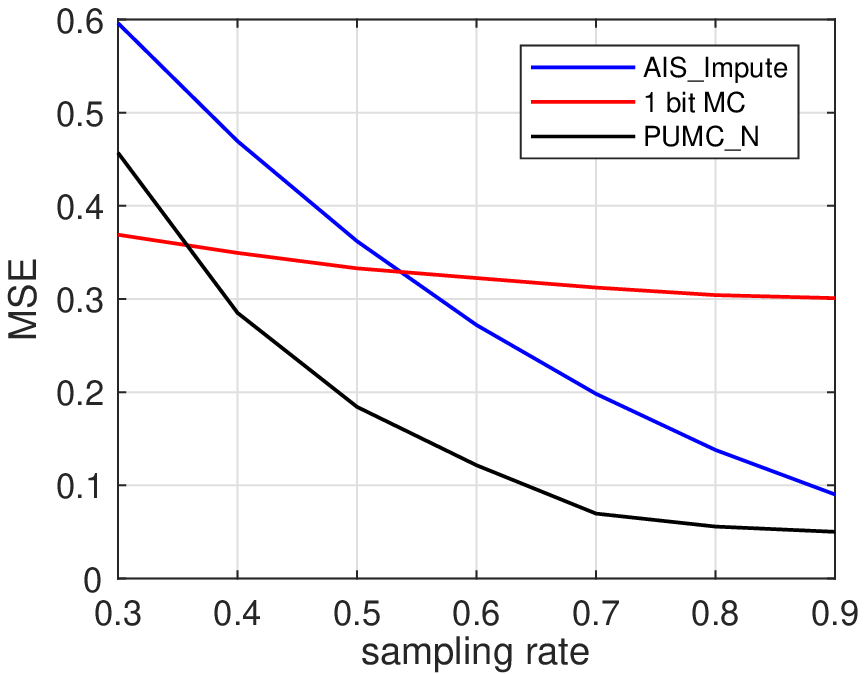}
		\label{subfig11}}
	\caption{Performance comparison of the recommender system methods. (a) Testing MSE vs sampling rate on the \textit{MovieLens} data set. (b) Testing MSE vs sampling rate on the \textit{FlimTrust} data set. (c) Testing MSE vs sampling rate on the \textit{Douban} data set.}
	\label{Fig.4}
\end{figure*}
\begin{table}[h]
	\centering
	\caption{Summary of Recommender system data sets.}
	\label{table5}
	\setlength{\tabcolsep}{5mm}{
		\begin{tabular}{@{}llll@{}}
			\toprule 
			& \#users & \#items & \#ratings \\
			\midrule
			\textit{MovieLens-100K} & 943     & 1,682   & 100,000   \\
			\textit{FlimTrust}      & 1,508   & 2,071   & 35,497    \\
			\textit{Douban}         & 3,000   & 3,000   & 136,891   \\
			\bottomrule
	\end{tabular}}
\end{table}

Results are shown in Fig. \ref{Fig.4}. Each point in the figure is the average across ten replicate experiments. Moreover, Table \ref{table6} shows the performance of the mentioned methods on three data sets. From the above experiments, LSP regularizer usually has better or comparable performance than the other two regularizers, thus we only use LSP regularizer here. It can be seen that in the three recommended algorithms, as long as the sampling rate is not less than 0.4, PUMC\_N will result in the lowest MSE in the least time. In addition, when the sampling rate is low, the performance of PUMC\_N needs to be improved. Moreover, the MSE vs time on the above three data sets is provided in Appendix B.

\section{Conclusion}
In this paper, we addressed the problem of binary matrix completion with nonconvex regularizers, where the observations consist only of positive entries. We proposed a novel PU matrix completion model \eqref{eq4} for tacking the task based on the commonly-used nonconvex regularizers and the $\omega$-weighted loss. In particular, the error bound for the model is derived to show the underlying matrix ${\mathbf{M}} \in {\left\{ {0,1} \right\}^{m \times n}}$ can be recovered accurately. Accordingly, we improved the proximal algorithm with two main acceleration strategies in nonconvex settings for solving \eqref{eq4}, and the convergence can also be guaranteed. The experiments on both synthetic and real-world data sets have verified the effectiveness of the proposed approach and validated the superiority over the state-of-the-art methods.

There still remain several directions for further work. From the experimental results, it can be seen that there is still room for further performance improvements at a low sampling rate. Besides, as in \cite{b24,b25}, a general nonuniform sampling distribution will be considered. In addition, to further speed up the proposed algorithm and apply it to massive data sets, we will focus on its distributed version.


\begin{thebibliography}{00}
	
	\bibitem{b1} 	E. J. Cand\`es and B. Recht, ``Exact matrix completion via convex optimization,'' \emph{Foundations of Computational Mathematics}, vol. 9, no. 6, pp. 717--772, 2009.
	
	\bibitem{b2} E. J. Candes and T. Tao, ``The power of convex relaxation: near-optimal matrix completion,'' \emph{IEEE Transactions on Information Theory}, vol. 56, no. 5, pp. 2053--2080, May 2010.
	
	\bibitem{b3} B. Recht, ``A simpler approach to matrix completion,'' \emph{ Journal of Machine Learning Research}, vol. 12, pp. 3413--3430, Dec. 2011.
	
	\bibitem{b4} A. Rohde and A. B. Tsybakov, ``Estimation of high-dimensional low-rank matrices,'' \emph{The Annals of Statistics}, vol. 39, no. 2, pp. 887--930, Apr. 2011.
	
	\bibitem{b5} R. H. Keshavan, A. Montanari, and S. Oh, ``Matrix completion from a few entries'' \emph{IEEE Transactions on Information Theory}, vol. 56, no. 6, pp. 2980--2998, Jun. 2010.
	
	\bibitem{b6} R. H. Keshavan, A. Montanari, and S. Oh, ``Matrix completion from noisy entries,'' \emph{Journal of Machine Learning Research}, vol. 11, no. 22, pp. 2057--2078, Aug. 2010.
	
	\bibitem{b7} T. T. Cai and W.-X. Zhou, ``Matrix completion via max-norm constrained optimization,''\emph{ Electronic Journal of Statistics}, vol. 10, no. 1, pp. 1493--1525, 2016.
	
	\bibitem{b8} Q. Sun, S. Xiang, and J.-P. Ye, ``Robust principal component analysis via capped norms,'' in \emph{Proceedings of the 19th ACM SIGKDD international conference on Knowledge discovery and data mining}, Chicago, Illinois, USA, 2013, pp. 311--319.
	
	\bibitem{b9} T. H. Oh, Y. W. Tai, J. C. Bazin, H. Kim, and I. S. Kweon, ``Partial sum minimization of singular values in robust PCA: algorithm and applications,'' \emph{IEEE Transactions on Pattern Analysis and Machine Intelligence}, vol. 38, no. 4, pp. 744--758, Apr. 2016.
	
	\bibitem{b10} Z. Wang, M. J. Lai, Z.-S. Lu, W. Fan, H. Davulcu, and J.-P Ye, ``Orthogonal rank-one matrix pursuit for low rank matrix completion,'' \emph{SIAM Journal on Scientific Computing}, vol. 37, no. 1, pp. A488--A514, Jan. 2015.
	
	\bibitem{b11} F. Xiao, W. Liu, Z. Li, L. Chen, and R. Wang, ``Noise-tolerant wireless sensor networks localization via multinorms regularized matrix completion,'' \emph{IEEE Transactions on Vehicular Technology}, vol. 67, no. 3, pp. 2409--2419, Mar. 2018.
	
	\bibitem{b12} C.-S Liu, H. Shan, and B. Wang, ``Wireless sensor network localization via matrix completion based on bregman divergence,'' \emph{Sensors}, vol. 18, no. 9, pp. 2974--2991, Sep. 2018.
	
	\bibitem{b13} C. Lu, J. Tang, S. Yan, and Z. Lin, ``Nonconvex nonsmooth low rank minimization via iteratively reweighted nuclear norm,'' \emph{IEEE Transactions on Image Processing}, vol. 25, no. 2, pp. 829--839, Feb. 2016.
	
	\bibitem{b14} S. Gu, Q. Xie, D. Meng, W. Zuo, X. Feng, and L. Zhang, ``Weighted nuclear norm minimization and its applications to low level vision,'' \emph{International Journal of Computer Vision}, vol. 121, no. 2, pp. 183--208, Jan. 2017.
	
	\bibitem{b15} R. Pech, D. Hao, L. Pan, H. Cheng, and T. Zhou, ``Link prediction via matrix completion,'' \emph{Europhysics Letters}, vol. 117, no. 3, pp. 38002, Mar. 2017.
	
	\bibitem{b16} M. A. Davenport, Y. Plan, E. van den Berg, and M. Wootters, ``1-bit matrix completion,'' \emph{Information and Inference}, vol. 3, no. 3, pp. 189--223, Sep. 2014.
	
	\bibitem{b17} C. Elkan and K. Noto, ``Learning classifiers from only positive and unlabeled data,'' in \emph{Proceeding of the 14th ACM SIGKDD international conference on Knowledge discovery and data mining}, Las Vegas, Nevada, USA, 2008, pp. 213--22.
	
	\bibitem{b18} C. J. Hsieh , N. Natarajan , and I. S. Dhillon, ``PU learning for matrix completion,'' \emph{Journal of Machine Learning Research},  vol 37, pp.2445--2453, Jul. 2015.
	
	\bibitem{b19} M. Hayashi , T. Sakai, and M. Sugiyama, ``Binary matrix completion using unobserved entries,''  \emph{arXiv}, 2018. [Online]. Available: \underline {https://arxiv.org/pdf/1803.04663.pdf} 
	
	\bibitem{b20} C. Lu, C. Zhu, C. Xu, S. Yan, and Z. Lin, ``Generalized singular value thresholding,'' in \emph{Proceedings of the 29th AAAI Conference on Artificial Intelligence}, Austin Texas, USA, 2015, pp. 1805--1811.
	
	\bibitem{b21} E. J. Cand\`es, M. B. Wakin, and S. P. Boyd, ``Enhancing sparsity by reweighted $l_1$ minimization,'' \emph{Journal of Fourier Analysis and Applications}, vol. 14, no. 5--6, pp. 877--905, Dec. 2008.
	
	\bibitem{b22} T. Zhang, ``Analysis of multi-stage convex relaxation for sparse regularization,'' \emph{Journal of Machine Learning Research},  vol 11, no. 3,  pp.1081--1107, Mar. 2010.
	
	\bibitem{b23} Y. Hu, D. Zhang, J. Ye, X. Li, and X. He, ``Fast and accurate matrix completion via truncated nuclear norm regularization,'' \emph{IEEE Transactions on Pattern Analysis and Machine Intelligence}, vol. 35, no. 9, pp. 2117--2130, Sep. 2013.
	
	\bibitem{b24} O. Klopp, J. Lafond, \'E. Moulines, and J. Salmon, ``Adaptive multinomial matrix completion,'' \emph{Electronic Journal of Statistics},  vol 9, no. 2, pp.2950--2975, 2015.
	
	\bibitem{b25} T. T. Cai, and W.-X. Zhou. ``A max-norm constrained minimization approach to 1-bit matrix completion,'' \emph{Journal of Machine Learning Research},  vol 14, no. 10, pp.3619--3647, Dec. 2013.
	
	\bibitem{b26} J. Wang, J. Shen, and .H Xu. ``Social trust prediction via max-norm constrained 1-bit matrix completion,'' \emph{arXiv}, Apr. 2015. [Online]. Available: \underline {https://arxiv.org/pdf/1504.06394.pdf}
	
	\bibitem{b27} S. A. Bhaskar and A. Javanmard, ``1-bit matrix completion under exact low-rank constraint,'' in \emph{2015 49th Annual Conference on Information Sciences and Systems}, Baltimore, MD, USA, 2015, pp. 1--6.
	
	\bibitem{b28} R. Ni and Q. Gu. ``Optimal statistical and computational rates for one bit matrix completion,'' \emph{Journal of Machine Learning Research}, vol 51, pp.426--434, May 2016.
	
	\bibitem{b29} V. Cottet and P. Alquier, ``1-bit matrix completion: PAC-Bayesian analysis of a variational approximation,'' \emph{Machine Learning}, vol. 107, no. 3, pp. 579--603, Mar. 2018.
	
	\bibitem{b30} R. Kiryo, G. Niu, and M. Plessis, and Masashi Sugiyama, ``Positive-unlabeled learning with non-negative risk estimator,'' \emph{31st Conference on Neural Information Processing Systems}, Long Beach, CA, USA, 2017, pp. 1675--1685.
	
	\bibitem{b31} T. Sakai, M. C. Plessis, G,Niu and M. Sugiyama, ``Semi-supervised classification based on classification from positive and unlabeled data,'' in \emph{Proceedings of the 34th International Conference on Machine Learning}, Sydney, Australia, 2017, pp. 2998--3006.
	
	\bibitem{b32} Q. Yao , and J. T. Kwok . ``Accelerated inexact soft-impute for fast large-scale matrix completion.'' in \emph{Proceedings of the 24th International Conference on Artificial Intelligence}, Buenos Aires, Argentina, 2015, pp. 4002--4008.
	
	\bibitem{b33} J. Cai, E. J. Cand\`es, and Z. Shen, ``A singular value thresholding algorithm for matrix completion,'' \emph{SIAM Journal on Optimization}, vol. 20, no. 4, pp. 1956--1982, Jan. 2010.
	
	\bibitem{b34} Z. Lin, M. Chen, and Y. Ma, ``The augmented lagrange multiplier method for exact recovery of corrupted low-rank matrices,'' \emph{Mathematical Programming}, vol. 9, Sep. 2010.
	
	\bibitem{b35} X. Su, Y. Wang, X. Kang, and R. Tao, ``Nonconvex truncated nuclear norm minimization based on adaptive bisection method,'' \emph{IEEE Transactions on Circuits and Systems for Video Technology}, pp. 1--14, Oct. 2018.
	
	\bibitem{b36} J. Yao, D. Meng, Q. Zhao, W. Cao, and Z. Xu, ``Nonconvex-sparsity and nonlocal-smoothness based blind hyperspectral unmixing,'' \emph{IEEE Transactions on Image Processing}, pp. 1--16, Jan. 2019.
	
	\bibitem{b37} V. Sindhwani, S. S. Bucak, J. Hu, and A. Mojsilovic, ``One-class matrix completion with low-density factorizations,'' in \emph{ 2010 IEEE International Conference on Data Mining}, Sydney, Australia, 2010, pp. 1055--1060.
	
	\bibitem{b38} C. Scott, ``Calibrated asymmetric surrogate losses,'' \emph{Electronic Journal of Statistics}, vol. 6, no. 0, pp. 958--992, Jan. 2012.
	
	\bibitem{b39} N. Natarajan, I. S. Dhillon,  P. Ravikumar, and A. Tewari, ``Learning with Noisy Labels,'' in \emph{Proceedings of the 26th International Conference on Neural Information Processing Systems}, Lake tahoe, Nevada, 2013, pp. 1196--1204.
	
	\bibitem{b40} P. L. Combettes and V. R. Wajs, ``Signal recovery by proximal forward-backward splitting,'' \emph{Multiscale Modeling and Simulation}, vol. 4, no. 4, pp. 1168--1200, 2005.
	
	\bibitem{b41} Q. Yao, J. T. Kwok, T. Wang, and T. Liu, ``Large-scale low-rank matrix learning with nonconvex regularizers,'' \emph{IEEE Transactions on Pattern Analysis and Machine Intelligence}, pp. 1--16, Jul, 2018.
	
	\bibitem{b42} R. Mazumder, T. Hastie, and R. Tibshirani, ``Spectral regularization algorithms for learning large incomplete matrices,'' \emph{Journal of Machine Learning Research}, vol. 11, no. 3, pp. 2287--2322, Aug. 2019.
	
	\bibitem{b43} N. Halko, P. G. Martinsson, and J. A. Tropp, ``Finding structure with randomness: probabilistic algorithms for constructing approximate matrix decompositions,'' \emph{SIAM Review}, vol. 53, no. 2, pp. 217--288, Jan. 2011.
	
	\bibitem{b44} H. Li, and Z. Lin, ``Accelerated proximal gradient methods for nonconvex programming,'' in \emph{Proceedings of the 28th International Conference on Neural Information Processing Systems}, Montreal, Canada, 2015, pp. 379--387.
	
	\bibitem{b45} H. Attouch, J. Bolte, and B. F. Svaiter, ``Convergence of descent methods for semi-algebraic and tame problems: proximal algorithms, forward-backward splitting, and regularized Gauss-Seidel methods,'' \emph{Mathematical Programming}, vol. 137, no. 1--2, pp. 91--129, Feb. 2013.
	
	\bibitem{b46} S. Segarra, A. G. Marques, G. Mateos, and A. Ribeiro, ``Network topology inference from spectral templates,'' \emph{IEEE Transactions on Signal and Information Processing over Networks}, vol. 3, no. 3, pp. 467--483, Sep. 2017.
	
	\bibitem{b47} P.-H Gong, C.-S Zhang, Z.-S Lu, J.-H Huang, and J.-P Ye, ``A general iterative shrinkage and thresholding algorithm for non-convex regularized optimization problems,'' \emph{Journal of Machine Learning Research}, vol. 28, no. 2, pp. 37--45, Mar. 2013.
	
	\bibitem{b48} P. M. Gleiser and L. Danon, ``Community structure in Jazz,'' \emph{Advances in Complex Systems}, vol. 6, no. 4, pp. 565--573, Dec. 2003.
	
	\bibitem{b49} M. Gao, L. Chen, B. Li, and W. Liu, ``A link prediction algorithm based on low-rank matrix completion,'' \emph{Applied Intelligence}, vol. 48, no. 12, pp. 4531--4550, Dec. 2018.
	
	\bibitem{b50} D. Liben-Nowell and J. Kleinberg, ``The link-prediction problem for social networks,'' \emph{Journal of the American Society for Information Science and Technology}, vol. 58, no. 7, pp. 1019--1031, May 2007.
	
	\bibitem{b51} A. Chaintreau, P. Hui, J. Crowcroft, C. Diot, R. Gass, and J. Scott, ``Impact of human mobility on opportunistic forwarding algorithms,'' \emph{IEEE Transactions on Mobile Computing}, vol. 6, no. 6, pp. 606--620, Jun. 2007.
	
	\bibitem{b52} J. Leskovec, D. Huttenlocher, and J. Kleinberg, ``Signed networks in social media,'' in \emph{Proceedings of the 28th international conference on Human factors in computing systems}, Atlanta, Georgia, USA, 2010, pp. 1361--1370.
	
	\bibitem{b53} G. Guo, J. Zhang, and N. Yorke-Smith, ``A novel evidence-based bayesian similarity measure for recommender systems,'' \emph{ACM Transactions on the Web}, vol. 10, no. 2, pp. 1--30, May 2016.
	
	\bibitem{b54} F. Monti, M. Bronstein, and X. Bresson, ``Geometric matrix completion with recurrent multi-graph neural networks,'' in \emph{Neural Information Processing Systems}, 2017, pp. 3697--3707.
	
\end{thebibliography}
\end{document}